\documentclass{article}

\usepackage{arxiv}
\usepackage{amsmath}
\usepackage[utf8]{inputenc} 
\usepackage[T1]{fontenc}    
\usepackage{hyperref}       
\usepackage{url}            
\usepackage{booktabs}       
\usepackage{amsfonts}       
\usepackage{nicefrac}       
\usepackage{microtype}      
\usepackage{lipsum}		
\usepackage{graphicx}
\usepackage[numbers,sort&compress]{natbib}
\usepackage{doi}
\usepackage{multirow}
\usepackage{bm}  

\title{GITO: Graph-Informed Transformer Operator for Learning Complex Partial Differential Equations}

\newcommand*\samethanks[1][\value{footnote}]{\footnotemark[#1]}

\author{{\hspace{1mm}Milad Ramezankhani}\thanks{These authors contributed equally to this work.}\\
	Applied Research, Quantiphi\\
        Marlborough, MA 01752, USA\\
	\texttt{milad.ramezankhani@quantiphi.com} \\
        \And
        {\hspace{1mm}Janak M. Patel}\samethanks\\
        Applied Research, Quantiphi\\
        Marlborough, MA 01752, USA\\
	\texttt{janak.patel@quantiphi.com} \\
        \And
	{\hspace{1mm}Anirudh Deodhar} \\
	Applied Research, Quantiphi\\
        Marlborough, MA 01752, USA\\
	\texttt{anirudh.deodhar@quantiphi.com} \\
        \And
	{\hspace{1mm}Dagnachew Birru} \\
	Applied Research, Quantiphi\\
        Marlborough, MA 01752, USA\\
	\texttt{dagnachew.birru@quantiphi.com} \\
}

\renewcommand{\shorttitle}{Graph-informed transformer operator for learning complex PDEs}

\hypersetup{
pdftitle={A template for the arxiv style},
pdfsubject={q-bio.NC, q-bio.QM},
pdfauthor={David S.~Hippocampus, Elias D.~Striatum},
pdfkeywords={First keyword, Second keyword, More},
}

\begin{document}
\maketitle

\begin{abstract}
We present a novel graph-informed transformer operator (GITO) architecture for learning complex partial differential equation systems defined on irregular geometries and non-uniform meshes. GITO consists of two main modules: a hybrid graph transformer (HGT) and a transformer neural operator (TNO). 
HGT leverages a graph neural network (GNN) to encode local spatial relationships and a transformer to capture long-range dependencies. A self-attention fusion layer integrates the outputs of the GNN and transformer to enable more expressive feature learning on graph-structured data. TNO module employs linear-complexity cross-attention and self-attention layers to map encoded input functions to predictions at arbitrary query locations, ensuring discretization invariance and enabling zero-shot super-resolution across any mesh. Empirical results on benchmark PDE tasks demonstrate that GITO outperforms existing transformer-based neural operators, paving the way for efficient, mesh-agnostic surrogate solvers in engineering applications.
\end{abstract}


\section{Introduction}
Solving partial differential equations (PDEs) underpins a vast array of phenomena in engineering and the physical sciences, from fluid flow and heat transfer to fracture mechanics and structural deformation. Traditional numerical methods offer rigorous error bounds and adaptable frameworks, but they often incur substantial computational costs when applied to high-dimensional, nonlinear, or time-dependent problems \citep{olver2014introduction}. This computational burden can become prohibitive in real-time control and optimization tasks, motivating the search for surrogate models that deliver rapid yet accurate PDE solutions.

In recent years, deep neural network–based surrogates have emerged as a powerful alternative, demonstrating orders-of-magnitude speedups over classical solvers while maintaining competitive accuracy \citep{zhu2018bayesian, bhatnagar2019prediction}. These data-driven models can learn solution operators from precomputed simulation data, enabling instantaneous inference once trained. Physics-informed neural networks (PINNs) \citep{raissi2019physics} introduced a paradigm shift by embedding the governing PDE residual directly into the loss function, thus bypassing the need for labeled solution data. While PINNs have been successfully applied to a wide range of forward and inverse problems, each new setting of initial conditions, boundary values, or forcing terms requires retraining from scratch, constraining their applicability to a single PDE configuration \citep{chen2024gpt, ramezankhani2024sequential}.

Neural operators extend the concept of surrogate modeling by directly mappings infinite-dimensional input-output spaces, effectively learning solution operators for a family of PDEs. Foundational architectures such as DeepONet \citep{lu2021learning} and the Fourier Neural Operator (FNO) \citep{li2020fourier} show that a single model can generalize across varying PDE conditions and enable zero-shot super-resolution. Inspired by the success of the transformer architecture \citep{vaswani2017attention} in natural language processing and computer vision, recent works explored attention-based surrogate models to simulate physical systems. Typically, these models are trained on function samples defined over fixed discretization grids, which limits their ability to generalize across varying meshes \citep{cao2021choose, han2022predicting}. To address this, a new class of transformer-based neural operators has emerged, which enables super-resolution and discretization-invariant query of the output function \citep{li2022transformer, hao2023gnot, alkin2024universal}. They employ cross-attention to aggregate input features and predict outputs at arbitrary spatial/temporal coordinates, regardless of the underlying input grid.

\begin{figure*}[htbp]
    \centering
    \includegraphics[width=0.9\linewidth]{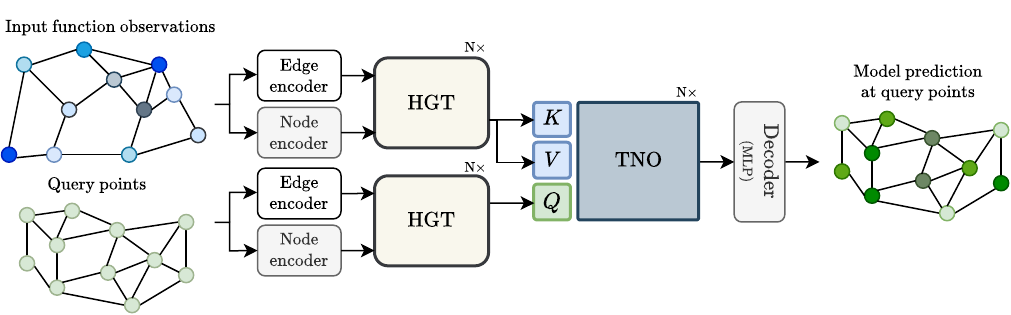} 
    \caption{Overall architecture of GITO. The input function and query points are first converted into graph representations and encoded via edge and node encoders. These encoded graphs are then processed by the hybrid graph transformer (HGT) module to learn informative relational features. The output representations from the HGT are used as key/value and query inputs to the transformer neural operator (TNO) module, which integrates contextual information from input function observations to enrich the query representations. Finally, an MLP decoder maps the query embeddings to real spatial coordinates.}
    \label{fig:full_model}
\end{figure*}

Despite these early successes, significant challenges remain in scaling transformer-based operators to realistic engineering applications. In particular, modeling systems with irregular geometries and non-uniform meshes demands more powerful mechanisms to capture complex interactions and dynamics among spatial nodes. To address these challenges, we propose a novel graph-informed transformer operator (GITO) architecture tailored for mesh-agnostic operator learning on arbitrary domains (Figure \ref{fig:full_model}). Our framework comprises two core modules: a hybrid graph transformer (HGT) and a transformer neural operator (TNO). HGT marries graph neural networks (GNNs) for modeling intricate local spatial relations with transformer layers for long-range, global dependencies, interleaving message-passing and self-attention via a dedicated fusion layer to produce expressive relational embeddings. Building on these embeddings, TNO applies cross-attention for discretization-invariant querying of the output domain, followed by self-attention to capture dependencies among enriched query embeddings. Our main contributions are: 1) a novel graph-transformer-based neural operator architecture that seamlessly integrates local and global feature learning on irregular meshes and geometries, and 2) superior performance on benchmark PDE tasks, outperforming existing transformer-based neural operators.

\section{Related work}
\textbf{Transformers as neural operators.} The attention mechanism has shown promise at modeling both spatial correlations and temporal dynamics in physical systems. Spatial attention layers aggregate information across nonlocal points, capturing structural patterns and long-range dependencies within the domain \citep{wu2024transolver, hao2023gnot, li2022transformer, bryutkin2024hamlet}. In the temporal setting, transformers learn state evolution over time without relying on recurrent architectures, often delegating spatial aggregation to other mechanisms such as GNNs \citep{alkin2024universal, han2022predicting, geneva2022transformers}. In addition, recent work has focused on developing novel transformer architectures to improve the scalability and effectiveness of modeling complex physical systems \citep{fonseca2023continuous, li2023scalable, chen2024positional}. Our method captures the spatial structures via linear-complexity attention mechanisms by leveraging the proposed HGT and TNO modules.

\textbf{Graphs as neural PDE solvers.} GNNs have been explored as mesh-agnostic PDE solvers by representing spatial discretizations as graph vertices and leveraging message-passing to model local interactions \citep{brandstetter2022message, li2020neural}. Previous studies have demonstrated that GNNs can effectively model diverse physical phenomena ranging from fluid dynamics and deformable materials \citep{sanchez2018graph} to global-scale weather forecasting \citep{lam2023learning}. Recently, transformer-inspired architectures have been applied to graph-based operator learning to more effectively handle arbitrary geometries and boundary conditions \citep{bryutkin2024hamlet}. In parallel, latent-space compression via graph encodings has enabled efficient dynamics propagation and scalable temporal rollouts \citep{alkin2024universal, han2022predicting}.

\section{Methodology}
\begin{figure}[htb]
    \centering
    \includegraphics[width=0.4\linewidth]{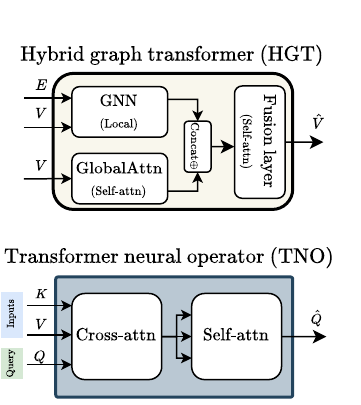} 
    \caption{(Top) The hybrid graph transformer (HGT) module consists of a GNN layer, a self-attention global layer, and a self-attention fusion layer that jointly learn graph-based representations. (Bottom) The transformer neural operator (TNO) module employs cross-attention and self-attention mechanisms to integrate and process representations of input functions and query points. For clarity, standard components such as layer normalization, residual connections, and feed-forward networks are omitted.}
    \label{fig:hgt_and_tno}
\end{figure}
\subsection{Graph construction and feature encoding}

We represent both the input function and query points as separate graphs $\mathcal{G}=(\mathcal{V}, \mathcal{E})$, where each node $i\in \mathcal{V}$ corresponds to a spatial location (e.g., a mesh cell or a query point) and each edge $(i,j)\in \mathcal{E}$ connects node $i$ either to its $k$ nearest neighbors or to nodes within a specified Euclidean radius. The value of $k$ and radius are considered as model hyperparameters (subsubsection \ref{knn}). Each node feature vector \( V_i \) includes the spatial coordinates \( \mathbf{x}_i \). For nodes corresponding to the input function, the observed field value \( \mathbf{u}_i \) is concatenated to the node features. Edge features $\mathbf{E}_{ij}$ comprise relative displacements $(\mathbf{x}_i - \mathbf{x}_j)$, Euclidean distances $|\mathbf{x}_i - \mathbf{x}_j|$, and, in case of input function graphs, differences in solution values between connected nodes $\mathbf{u}_i - \mathbf{u}_j$ \citep{brandstetter2022message}. Both node and edge features are passed through dedicated MLP-based encoders to generate initial embeddings, which are then fed into the HGT layers for subsequent representation learning.
\subsection{Hybrid graph transformer (HGT) module}
Despite their strengths, GNNs suffer from fundamental limitations due to sparse message passing, notably over-smoothing \citep{oono2019graph} and over-squashing \citep{alon2020bottleneck}. Graph transformers (GTs) \citep{dwivedi2020generalization, ying2021transformers, mialon2021graphit} address these shortcomings by allowing nodes to attend to all others in the graph; however, they often overlook edge features, hindering accurate representation learning. Hybrid architectures such as GPS Graph \citep{rampavsek2022recipe} and Exphormer \citep{shirzad2023exphormer} combine GNN and transformer layers to overcome these challenges: the GNN component captures local interactions and integrates edge information, while the transformer module models long-range and global dependencies and mitigates over-smoothing and over-squashing. Following this paradigm, we employ a GNN layer (\texttt{GNN}) alongside a linear self-attention module (\texttt{GlobalAttn}) to learn graph dynamics and introduce a fusion layer (\texttt{Fusion}) that applies self-attention to interleave local neighborhood aggregation with global attention, resulting in richer and more expressive graph representations (Figure \ref{fig:hgt_and_tno}). In the HGT module, node representations are updated by concatenating the outputs of the \texttt{GNN} and \texttt{GlobalAttn} layers, followed by processing the combined embedding through the \texttt{Fusion} layer:
\begin{align}
V_G, E &= \texttt{GNN}(V, E) \tag{1} \\
V_T &= \texttt{GlobalAttn}(V) \tag{2} \\
\hat{V} &= \texttt{Fusion}\left(V_G \oplus V_T\right) \tag{3}.
\end{align}
The modularity of the hybrid graph transformer enables seamless integration of diverse GNN architectures and transformer modules, allowing the model to be tailored to specific application requirements.
\subsection{Transformer neural operator (TNO) module}
To empower zero-shot super-resolution and fully decouple input and output representations, we integrate a cross-attention layer capable of querying the output domain at arbitrary spatial locations (Figure \ref{fig:hgt_and_tno}). This design parallels the branch and trunk networks in the DeepONet \citep{li2020neural}, seamlessly fusing input function embeddings with output queries to achieve discretization-invariant evaluation, regardless of the underlying input mesh \citep{li2022transformer}. The cross-attention layer takes as input the query embeddings and the input function representations generated by the HGT modules. The cross-attention enriches the query embeddings with the information from the input functions. A subsequent self-attention module then captures interactions and dependencies among the enriched query points. Finally, an MLP decoder translates the resulting embeddings into the target physical output values.

\subsection{Model implementation details}
To efficiently learn operators for large-scale physical systems with numerous input and query locations, we adopt the linear-complexity attention mechanism proposed by \citet{hao2023gnot}. Similar to Fourier and Galerkin attention mechanisms \citep{cao2021choose}, this approach can capture complex dynamics while avoiding the quadratic computational cost of softmax-based attention. We adopt a “Norm-Attn-MLP-Norm” with residual connections for all attention layers. To handle cases with multiple input functions, we use a dedicated encoder for each input function. These encoded representations are then processed by the cross-attention module in TNO, specifically designed to handle multiple key-value (K/V) combinations, enabling efficient interaction across heterogeneous inputs. We incorporate a mixture of experts module following each attention mechanism. The gating network assigns weights to the experts based on the spatial location of the query points, effectively promoting a form of \textit{soft} domain decomposition, which has been shown to enhance the learning of physical systems in prior work \citep{chalapathi2024scaling, hao2023gnot}.  In the HGT module, we use the Graph Attention Network (GATv2) \citep{brody2021attentive} as the GNN layer and apply the same linear-complexity attention mechanism as in TNO for both the global and fusion layers. The graph construction strategies are detailed in subsubsection~\ref{knn}. In this work, we choose to use the HGT module only for query points for learning more expressive relational features.
\section{Experimental results}
\subsection{Datasets and baseline models}
\textbf{Datasets.} To evaluate GITO’s scalability on complex geometries, we test it on three challenging datasets: Navier-Stokes \citep{hao2023gnot}, Heat Conduction \citep{hao2023gnot}, and Airfoil \citep{li2022fourier}. A brief overview of the datasets is provided below, with detailed descriptions available in Appendix~\ref{Appendix:dataset}:
\begin{itemize}
    \item \textbf{2D steady-state Navier-Stokes (NS)}: This dataset involves steady 2D fluid flow governed by Navier-Stokes equations in a rectangular domain with varying cavity positions (Figure \ref{fig:mesh}.a). The goal is to predict velocity components $u$, $v$, and pressure $p$ from the input mesh.
    \item \textbf{Multilayer 2D Heat Conduction (Heat)}: This dataset models heat conduction in composite media with multiple boundary shapes and spatially varying boundary conditions (Figure \ref{fig:mesh}.b). The task is to predict temperature $T$ from multiple input functions.
    \item \textbf{Airfoil}: This dataset involves the Mach number $M$ distribution over different airfoil shapes (Figure \ref{fig:airfoilmesh}). The task is to predict $M$ from the input mesh and the geometry of the airfoil.
\end{itemize}

\textbf{Baseline Models.} We benchmark our model against both conventional neural operator architectures, including FNO \citep{li2020fourier}, Geo-FNO \citep{li2022fourier}, and MIONet \citep{jin2022mionet}, as well as recently developed transformer-based operators, namely, GNOT \citep{hao2023gnot}, Galerkin Transformer (GKT) \citep{cao2021choose}, and OFormer \citep{li2022transformer}. To ensure a fair comparison, we re-implement GNOT and evaluate it under the same experimental settings as our model, using a comparable or slightly larger number of parameters (Appendix \ref{Appendix:dataset}). For the NS and Heat datasets, we directly report the baseline performances from \citet{hao2023gnot}, while for the Airfoil dataset, we refer to \citet{wu2024transolver}.

\subsection{Results} 
Table~\ref{tab:error} summarizes the mean relative $L^2$ error across all test datasets for the compared models, with lower values indicating higher prediction accuracy. The relative $L^2$ error is defined as $\frac{|\hat{y} - y|_2}{|y|_2}$, where $\hat{y}$ is the model prediction and $y$ is the ground truth. This metric provides a normalized measure of prediction accuracy that is consistent across datasets with varying magnitudes. Detailed configurations and hyperparameter settings are provided in Appendix~\ref{Appendix:dataset}. GITO consistently achieves the lowest error across all datasets and variables, outperforming existing neural operator baselines. This includes both conventional architectures such as FNO~\citep{li2020fourier}, Geo-FNO~\citep{li2022fourier}, and MIONet~\citep{jin2022mionet}, as well as transformer-based models like GNOT~\citep{hao2023gnot}, GKT~\citep{cao2021choose}, and OFormer~\citep{li2022transformer}. To ensure a fair comparison, we re-trained GNOT using a reduced model size that matches GITO's parameter count. Although FNO slightly outperforms GNOT for the $p$ variable on the NS dataset, GITO surpasses both, demonstrating superior generalization capabilities. For the Airfoil dataset, while the GKT model achieves the best baseline performance on this dataset, our proposed GITO model delivers more than 46\% improvement in prediction accuracy over GKT, highlighting its effectiveness in modeling complex geometries.

\begin{table*}[htb]
\centering
\begin{tabular}{@{}c|c|c|c|c|c|c|c|c|c@{}}
\toprule
Dataset & Subset & MIONet  & FNO           & GKT     & Geo-FNO & OFormer & GNOT          & \textbf{GITO (Ours)} & Improvement \\ \midrule
\multirow{3}{*}{NS} & $u$ & 2.74e-2 & 6.56e-2 & 1.52e-2 & 1.41e-2 & 2.33e-2 & {\underline{1.05e-2}} & \textbf{8.19e-3} & 22 \% \\
        & $v$    & 5.51e-2 & 1.15e-1       & 3.15e-2 & 2.98e-2 & 4.83e-2 & {\underline{2.33e-2}} & \textbf{2.02e-2}     & 13.3 \%     \\
        & $p$    & 2.74e-2 & {\underline{1.11e-2}} & 1.59e-2 & 1.62e-2 & 2.43e-2 & 1.23e-2       & \textbf{1.07e-2}     & 3.6 \%      \\ \midrule
Heat    & $T$    & 1.74e-1 & --            & --      & --      & --      & {\underline{5.42e-2}} & \textbf{4.49e-2}     & 17.2 \%     \\ \midrule
Airfoil & $M$    &   --      & --            & \underline{1.18e-2} & 1.38e-2 & 1.83e-2 &    1.80e-2           &         \textbf{6.29e-3}              &    46.7 \%         \\ \bottomrule
\end{tabular}
\caption{Comparison of GITO with existing operator learning methods on the NS, Heat, and Airfoil datasets. The metric used for this comparison is relative $L^2$ error, with lower scores indicating better performance. The top \textbf{first} and \underline{second} best results are highlighted. For the NS and Heat datasets, baseline results (except GNOT) are taken directly from \citet{hao2023gnot}, while for the Airfoil dataset, they are taken from \citet{wu2024transolver}. For a fair comparison, we trained a smaller GNOT model to match GITO's model size (see Appendix \ref{Appendix:dataset} for details.)}
\label{tab:error}
\vspace{-1.5ex}
\end{table*}

\begin{figure}
    \centering
    \includegraphics[width=1\textwidth]{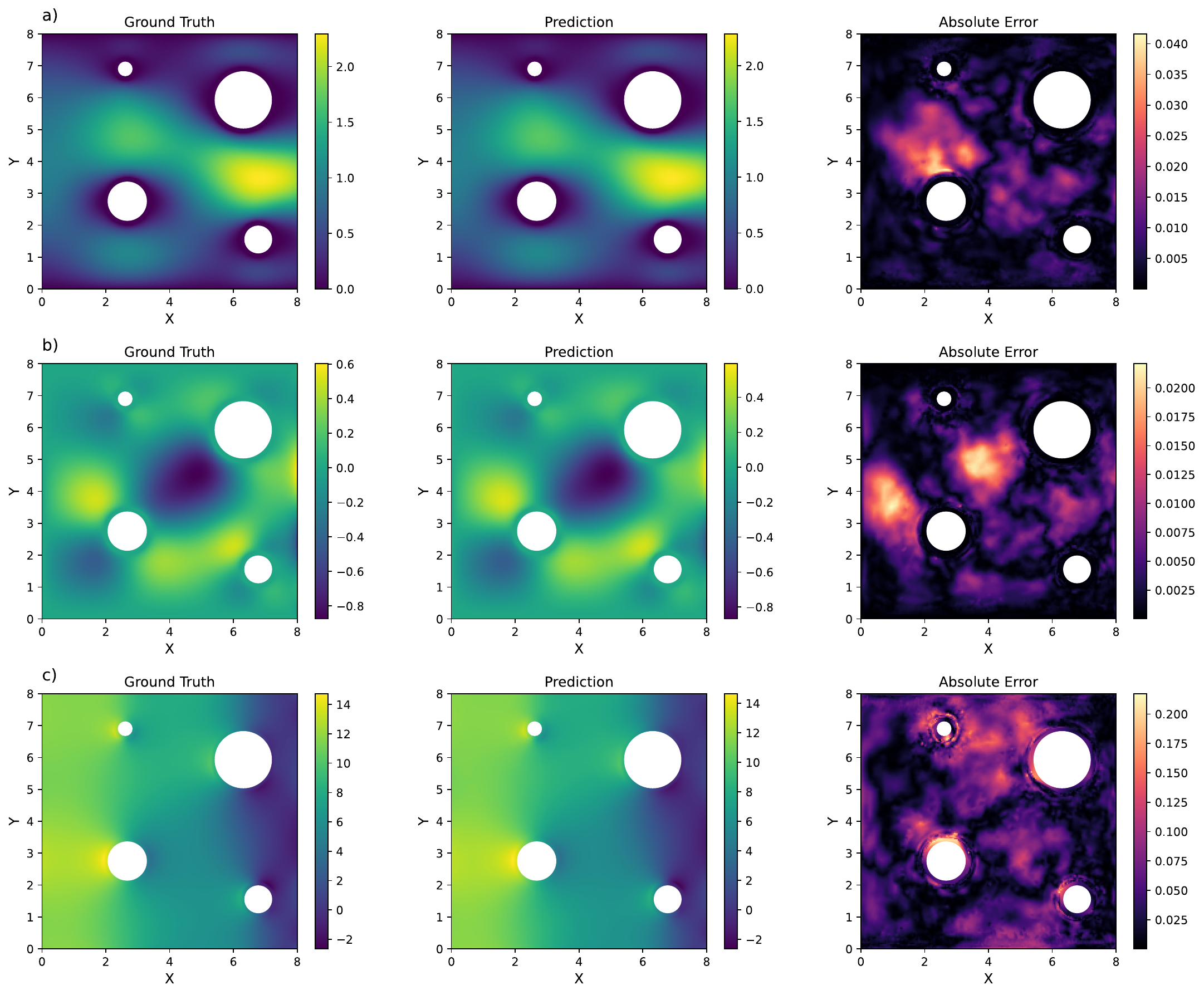}
    \caption{Comparison of GITO’s predictions against ground truth, and the corresponding absolute error plots for a test sample from the NS dataset: a) velocity component $u$, b) velocity component $v$, and c) pressure $p$.}
    \label{fig:ns2d_result}
\end{figure}

\begin{figure}
    \centering
    \includegraphics[width=1\textwidth]{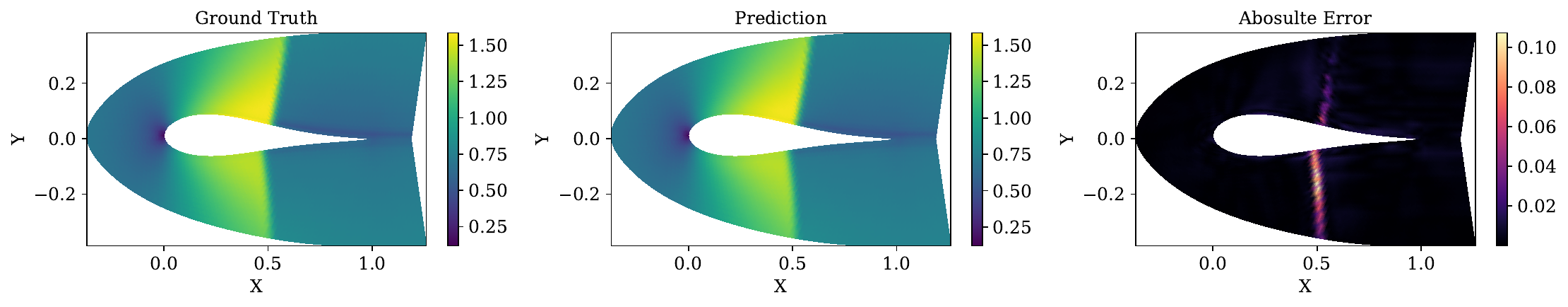}
    \caption{Comparison between GITO’s predicted Mach number and the ground truth for a representative test sample from the Airfoil dataset. The left and middle panels show the ground truth and predicted fields, respectively, while the right panel presents the absolute error.}
    \label{fig:airfoil_result}
\end{figure}

Figures~\ref{fig:ns2d_result} and~\ref{fig:airfoil_result} further illustrate GITO’s generalization capability. In particular, Figure~\ref{fig:ns2d_result} presents a qualitative comparison of velocity components $(u, v)$ and pressure $(p)$ predicted by GITO against the ground truth for a sample from the NS dataset. The corresponding absolute error plots reveal the spatial distribution of prediction inaccuracies. Likewise, Figure~\ref{fig:airfoil_result} demonstrates GITO's predictions of the Mach number field for a representative sample from the Airfoil dataset. The predicted field closely matches the ground truth, and the absolute error plot indicates minimal residual error. These findings demonstrate GITO’s generality and efficacy in handling both complex geometries (NS and Airfoil dataset) and multi-input settings (Heat dataset), establishing it as a versatile, high-performance surrogate for diverse scientific and engineering applications.

\subsection{Ablation studies}
Beyond overall performance, we conducted extensive ablation studies to assess the effect of key architectural components and design choices.

\subsubsection{Impact of fusion layer} \label{fusion}
To demonstrate the effect of the fusion layer on model accuracy, we conducted experiments on the NS dataset using identical hyperparameters, except for the hidden size. In this experiment, instead of concatenating the outputs of the GNN and self-attention modules, we summed them and passed the result through an MLP (similar to GPS Graph \citep{rampavsek2022recipe}). Accordingly, the model without the fusion layer was trained with a hidden size of 192 (twice that of GITO) to match the dimensionality of the fused outputs. The number of model parameters and the $L^2$ relative error on the NS dataset are reported in Table~\ref{tab:fusion}. It is evident that the model without the fusion layer exhibits degraded accuracy, despite having a larger number of parameters. This clearly demonstrates the effectiveness of the fusion mechanism in enabling more expressive feature interactions between the GNN and self-attention outputs, as opposed to the limited representational capacity of simple element-wise summation.
\begin{table*}[h]
\centering
\begin{tabular}{c|c|c|c}
\toprule
\multicolumn{2}{c|}{\textbf{Configuration$\rightarrow$}} & \textbf{GITO w/ Fusion} & \textbf{GITO w/o Fusion} \\ \midrule
\multirow{3}{*}{\textbf{Relative $L^2$ Error}} 
  & $u$ & 8.42e-3 &  1.00e-2 \\
  & $v$ & 2.06e-2 & 2.44e-2 \\
  & $p$ & 1.12e-2 &  1.61e-2 \\ \midrule
\multicolumn{2}{c|}{\textbf{Model Parameters (M)}} & 4.75 & 5.35 \\
\bottomrule
\end{tabular}
\caption{Ablation study comparing GITO with and without the fusion layer on the NS dataset. The fusion layer combines outputs from the GNN and self-attention paths. Reported values are relative $L^2$ errors; lower is better.} 
\label{tab:fusion}
\vspace{-1.5ex}
\end{table*}
\subsubsection{Effect of graph construction strategies} \label{knn}
We conduct an ablation study to evaluate the impact of graph construction methods on the accuracy and computational efficiency of the proposed GITO model. Specifically, we compare two widely used strategies: \textbf{K-nearest neighbors (KNN)} and \textbf{radius-based (circular)} graphs \cite{qi2017pointnet++, bryutkin2024hamlet}. In the KNN strategy, each node is connected to its \( k \) nearest neighbors based on spatial proximity. In the circular strategy, nodes are connected to all other nodes within a fixed radius, forming edges only if the pairwise distance falls below the specified threshold.

\textbf{NS Dataset.}  
As shown in Table~\ref{tab:graph_ns}, increasing the number of neighbors in KNN (from 4 to 16) consistently reduces the relative \(L^2\) error, indicating that denser local connectivity enables better modeling of spatial dependencies. However, this improvement comes at the cost of a significant increase in the number of graph edges (from 41k to 164k), which leads to higher memory usage and computational time during training and inference. On the other hand, the circular strategy with a radius of 0.0525 achieves the lowest error (3.91e-2) while maintaining a moderate edge count (16k). This suggests that, with careful tuning, the radius-based approach can capture only the most relevant local interactions and avoid unnecessary edges, providing a better balance between accuracy and efficiency. Larger radius, such as 0.067, may include irrelevant distant nodes, while smaller radius (e.g., 0.04) risk missing essential local connections—both resulting in slightly degraded performance.
\begin{table*}[htbp]
\centering
\begin{tabular}{@{}c|ccc|ccc@{}}
\toprule
\multirow{2}{*}{\textbf{\begin{tabular}[c]{@{}c@{}}Strategy for graph \\ construction\end{tabular}}} &
  \multicolumn{3}{c|}{\textbf{KNN (Number of neighbors)}} &
  \multicolumn{3}{c}{\textbf{Circular (Radius)}} \\ \cmidrule(l){2-7} 
 &
  \multicolumn{1}{c|}{\textbf{4}} &
  \multicolumn{1}{c|}{\textbf{8}} &
  \textbf{16} &
  \multicolumn{1}{c|}{\textbf{0.04}} &
  \multicolumn{1}{c|}{\textbf{0.0525}} &
  \textbf{0.067} \\ \midrule
\textbf{Relative $L^2$ error (total)} &
  \multicolumn{1}{c|}{4.20e-2} &
  \multicolumn{1}{c|}{4.33e-3} &
4.17e-2 &
  \multicolumn{1}{c|}{4.06e-2} &
  \multicolumn{1}{c|}{3.91e-2} &
  3.93e-2 \\ \midrule
\textbf{Number of Edges} &
  \multicolumn{1}{c|}{41k} &
  \multicolumn{1}{c|}{82k} &
  164k &
  \multicolumn{1}{c|}{8k} &
  \multicolumn{1}{c|}{16k} &
  44k \\ \bottomrule
\end{tabular}%
\caption{Ablation study on the effect of graph construction strategies-KNN (with varying number of neighbors) and Circular (with varying radius)-on model accuracy for the NS dataset. The table reports the sum of relative $L^2$ errors across all three variables (lower is better). The number of edges is reported approximately in thousands.} 
\label{tab:graph_ns}
\vspace{-1.5ex}
\end{table*}

\textbf{Heat Dataset.}  
The trends differ for the Heat dataset (Table~\ref{tab:graph_heat}), which features more sparsely distributed query points and different material properties. Here, increasing the KNN count beyond 8 does not yield consistent improvements and, in fact, degrades performance. For instance, using 16 neighbors increases the error to 4.75e-2 compared to 4.61e-2 with 8 neighbors and 4.49e-2 with only 4 neighbors. This is likely because higher KNN values may force connections to spatially distant and physically irrelevant nodes, misleading the model in heterogeneous material settings. Similarly, larger radius in the circular graph (e.g., 0.9) also lead to performance drops due to excessive inclusion of unrelated nodes. The best performance is observed with a small radius (0.25), which maintains sparse yet contextually meaningful connectivity. These results emphasize the importance of tailoring graph construction strategies to the underlying spatial structure and physical properties of the dataset.

\textbf{Airfoil dataset.} For the Airfoil dataset, we used a KNN graph with 16 neighbors to ensure consistent connectivity across the large, sparsely discretized domain. Radius-based graphs either led to disconnected nodes in sparse regions or overly dense connections in clustered areas. KNN provided a balanced neighborhood structure, improving information flow and resulting in better prediction accuracy for the Mach field.

\begin{table*}[htbp]
\centering
\begin{tabular}{@{}cl|ccc|ccc@{}}
\toprule
\multicolumn{2}{c|}{\multirow{2}{*}{\textbf{\begin{tabular}[c]{@{}c@{}}Strategy for graph \\ construction\end{tabular}}}} &
  \multicolumn{3}{c|}{\textbf{KNN (Number of neighbors)}} &
  \multicolumn{3}{c}{\textbf{Circular (Radius)}} \\ \cmidrule(l){3-8} 
\multicolumn{2}{c|}{} &
  \multicolumn{1}{c|}{\textbf{4}} &
  \multicolumn{1}{c|}{\textbf{8}} &
  \textbf{16} &
  \multicolumn{1}{c|}{\textbf{0.25}} &
  \multicolumn{1}{c|}{\textbf{0.4}} &
  \textbf{0.9} \\ \midrule
\multicolumn{2}{c|}{\textbf{Relative $L^2$ error}} &
  \multicolumn{1}{c|}{4.49e-2} &
  \multicolumn{1}{c|}{4.61e-2} &
  4.75e-2 &
  \multicolumn{1}{c|}{4.64e-2} &
  \multicolumn{1}{c|}{4.72e-2} &
  4.75e-2 \\ \midrule
\multicolumn{2}{c|}{\textbf{Number of Edges}} &
  \multicolumn{1}{c|}{8k} &
  \multicolumn{1}{c|}{16k} &
  33k &
  \multicolumn{1}{c|}{2.5k} &
  \multicolumn{1}{c|}{8k} &
  36k \\ \bottomrule
\end{tabular}
\caption{Ablation study on the effect of graph construction strategies on model performance for temperature prediction in the Heat dataset. Reported values are relative $L^2$ errors; lower is better. The number of edges is reported approximately in thousands.} 
\label{tab:graph_heat}
\vspace{-1.5ex}
\end{table*}

Overall, the ablation studies demonstrate that the choice of graph construction strategy significantly affects both the accuracy and computational efficiency of the model. While KNN provides a simple and adaptive structure, circular graphs—when carefully tuned—offer a more interpretable and controllable connectivity pattern. For datasets with dense spatial coverage (like NS), moderate-radius circular graphs are preferable, while for sparse or heterogeneous domains (like Heat), lower connectivity thresholds help prevent overfitting to irrelevant neighbors. Ultimately, the best graph construction strategy varies depending on the specific characteristics of the problem domain.

\section{Conclusion}
In this work, we introduced GITO, the Graph-Informed Transformer Operator, a novel architecture that unifies graph neural networks with transformer attention to learn mesh-agnostic PDE solution operators for arbitrary geometries. By combining hybrid message-passing, discretization-invariant cross-attention, and scalable linear-complexity attention mechanisms, GITO delivers zero-shot super-resolution and outperforms existing transformer-based operators across diverse benchmarks. These results underscore GITO’s promise as an accurate and efficient surrogate model for complex engineering applications.

\bibliographystyle{unsrtnat}
\bibliography{references}  

\begin{thebibliography}{37}
\providecommand{\natexlab}[1]{#1}
\providecommand{\url}[1]{\texttt{#1}}
\expandafter\ifx\csname urlstyle\endcsname\relax
  \providecommand{\doi}[1]{doi: #1}\else
  \providecommand{\doi}{doi: \begingroup \urlstyle{rm}\Url}\fi

\bibitem[Olver et~al.(2014)]{olver2014introduction}
Peter~J Olver et~al.
\newblock \emph{Introduction to partial differential equations}, volume~1.
\newblock Springer, 2014.

\bibitem[Zhu and Zabaras(2018)]{zhu2018bayesian}
Yinhao Zhu and Nicholas Zabaras.
\newblock Bayesian deep convolutional encoder--decoder networks for surrogate modeling and uncertainty quantification.
\newblock \emph{Journal of Computational Physics}, 366:\penalty0 415--447, 2018.

\bibitem[Bhatnagar et~al.(2019)Bhatnagar, Afshar, Pan, Duraisamy, and Kaushik]{bhatnagar2019prediction}
Saakaar Bhatnagar, Yaser Afshar, Shaowu Pan, Karthik Duraisamy, and Shailendra Kaushik.
\newblock Prediction of aerodynamic flow fields using convolutional neural networks.
\newblock \emph{Computational Mechanics}, 64:\penalty0 525--545, 2019.

\bibitem[Raissi et~al.(2019)Raissi, Perdikaris, and Karniadakis]{raissi2019physics}
Maziar Raissi, Paris Perdikaris, and George~E Karniadakis.
\newblock Physics-informed neural networks: A deep learning framework for solving forward and inverse problems involving nonlinear partial differential equations.
\newblock \emph{Journal of Computational physics}, 378:\penalty0 686--707, 2019.

\bibitem[Chen and Koohy(2024)]{chen2024gpt}
Yanlai Chen and Shawn Koohy.
\newblock Gpt-pinn: Generative pre-trained physics-informed neural networks toward non-intrusive meta-learning of parametric pdes.
\newblock \emph{Finite Elements in Analysis and Design}, 228:\penalty0 104047, 2024.

\bibitem[Ramezankhani and Milani(2024)]{ramezankhani2024sequential}
Milad Ramezankhani and Abbas~S Milani.
\newblock A sequential meta-transfer (smt) learning to combat complexities of physics-informed neural networks: Application to composites autoclave processing.
\newblock \emph{Composites Part B: Engineering}, 283:\penalty0 111597, 2024.

\bibitem[Lu et~al.(2021)Lu, Jin, Pang, Zhang, and Karniadakis]{lu2021learning}
Lu~Lu, Pengzhan Jin, Guofei Pang, Zhongqiang Zhang, and George~Em Karniadakis.
\newblock Learning nonlinear operators via deeponet based on the universal approximation theorem of operators.
\newblock \emph{Nature machine intelligence}, 3\penalty0 (3):\penalty0 218--229, 2021.

\bibitem[Li et~al.(2020{\natexlab{a}})Li, Kovachki, Azizzadenesheli, Liu, Bhattacharya, Stuart, and Anandkumar]{li2020fourier}
Zongyi Li, Nikola Kovachki, Kamyar Azizzadenesheli, Burigede Liu, Kaushik Bhattacharya, Andrew Stuart, and Anima Anandkumar.
\newblock Fourier neural operator for parametric partial differential equations.
\newblock \emph{arXiv preprint arXiv:2010.08895}, 2020{\natexlab{a}}.

\bibitem[Vaswani et~al.(2017)Vaswani, Shazeer, Parmar, Uszkoreit, Jones, Gomez, Kaiser, and Polosukhin]{vaswani2017attention}
Ashish Vaswani, Noam Shazeer, Niki Parmar, Jakob Uszkoreit, Llion Jones, Aidan~N Gomez, {\L}ukasz Kaiser, and Illia Polosukhin.
\newblock Attention is all you need.
\newblock \emph{Advances in neural information processing systems}, 30, 2017.

\bibitem[Cao(2021)]{cao2021choose}
Shuhao Cao.
\newblock Choose a transformer: Fourier or galerkin.
\newblock \emph{Advances in neural information processing systems}, 34:\penalty0 24924--24940, 2021.

\bibitem[Han et~al.(2022)Han, Gao, Pfaff, Wang, and Liu]{han2022predicting}
Xu~Han, Han Gao, Tobias Pfaff, Jian-Xun Wang, and Li-Ping Liu.
\newblock Predicting physics in mesh-reduced space with temporal attention.
\newblock \emph{arXiv preprint arXiv:2201.09113}, 2022.

\bibitem[Li et~al.(2022{\natexlab{a}})Li, Meidani, and Farimani]{li2022transformer}
Zijie Li, Kazem Meidani, and Amir~Barati Farimani.
\newblock Transformer for partial differential equations' operator learning.
\newblock \emph{arXiv preprint arXiv:2205.13671}, 2022{\natexlab{a}}.

\bibitem[Hao et~al.(2023)Hao, Wang, Su, Ying, Dong, Liu, Cheng, Song, and Zhu]{hao2023gnot}
Zhongkai Hao, Zhengyi Wang, Hang Su, Chengyang Ying, Yinpeng Dong, Songming Liu, Ze~Cheng, Jian Song, and Jun Zhu.
\newblock Gnot: A general neural operator transformer for operator learning.
\newblock In \emph{International Conference on Machine Learning}, pages 12556--12569. PMLR, 2023.

\bibitem[Alkin et~al.(2024)Alkin, F{\"u}rst, Schmid, Gruber, Holzleitner, and Brandstetter]{alkin2024universal}
Benedikt Alkin, Andreas F{\"u}rst, Simon Schmid, Lukas Gruber, Markus Holzleitner, and Johannes Brandstetter.
\newblock Universal physics transformers: A framework for efficiently scaling neural operators.
\newblock \emph{Advances in Neural Information Processing Systems}, 37:\penalty0 25152--25194, 2024.

\bibitem[Wu et~al.(2024)Wu, Luo, Wang, Wang, and Long]{wu2024transolver}
Haixu Wu, Huakun Luo, Haowen Wang, Jianmin Wang, and Mingsheng Long.
\newblock Transolver: A fast transformer solver for pdes on general geometries.
\newblock \emph{arXiv preprint arXiv:2402.02366}, 2024.

\bibitem[Bryutkin et~al.(2024)Bryutkin, Huang, Deng, Yang, Sch{\"o}nlieb, and Aviles-Rivero]{bryutkin2024hamlet}
Andrey Bryutkin, Jiahao Huang, Zhongying Deng, Guang Yang, Carola-Bibiane Sch{\"o}nlieb, and Angelica Aviles-Rivero.
\newblock Hamlet: Graph transformer neural operator for partial differential equations.
\newblock \emph{arXiv preprint arXiv:2402.03541}, 2024.

\bibitem[Geneva and Zabaras(2022)]{geneva2022transformers}
Nicholas Geneva and Nicholas Zabaras.
\newblock Transformers for modeling physical systems.
\newblock \emph{Neural Networks}, 146:\penalty0 272--289, 2022.

\bibitem[Fonseca et~al.(2023)Fonseca, Zappala, Caro, and Van~Dijk]{fonseca2023continuous}
Antonio H de~O Fonseca, Emanuele Zappala, Josue~Ortega Caro, and David Van~Dijk.
\newblock Continuous spatiotemporal transformers.
\newblock \emph{arXiv preprint arXiv:2301.13338}, 2023.

\bibitem[Li et~al.(2023)Li, Shu, and Barati~Farimani]{li2023scalable}
Zijie Li, Dule Shu, and Amir Barati~Farimani.
\newblock Scalable transformer for pde surrogate modeling.
\newblock \emph{Advances in Neural Information Processing Systems}, 36:\penalty0 28010--28039, 2023.

\bibitem[Chen and Wu(2024)]{chen2024positional}
Junfeng Chen and Kailiang Wu.
\newblock Positional knowledge is all you need: Position-induced transformer (pit) for operator learning.
\newblock \emph{arXiv preprint arXiv:2405.09285}, 2024.

\bibitem[Brandstetter et~al.(2022)Brandstetter, Worrall, and Welling]{brandstetter2022message}
Johannes Brandstetter, Daniel Worrall, and Max Welling.
\newblock Message passing neural pde solvers.
\newblock \emph{arXiv preprint arXiv:2202.03376}, 2022.

\bibitem[Li et~al.(2020{\natexlab{b}})Li, Kovachki, Azizzadenesheli, Liu, Bhattacharya, Stuart, and Anandkumar]{li2020neural}
Zongyi Li, Nikola Kovachki, Kamyar Azizzadenesheli, Burigede Liu, Kaushik Bhattacharya, Andrew Stuart, and Anima Anandkumar.
\newblock Neural operator: Graph kernel network for partial differential equations.
\newblock \emph{arXiv preprint arXiv:2003.03485}, 2020{\natexlab{b}}.

\bibitem[Sanchez-Gonzalez et~al.(2018)Sanchez-Gonzalez, Heess, Springenberg, Merel, Riedmiller, Hadsell, and Battaglia]{sanchez2018graph}
Alvaro Sanchez-Gonzalez, Nicolas Heess, Jost~Tobias Springenberg, Josh Merel, Martin Riedmiller, Raia Hadsell, and Peter Battaglia.
\newblock Graph networks as learnable physics engines for inference and control.
\newblock In \emph{International conference on machine learning}, pages 4470--4479. PMLR, 2018.

\bibitem[Lam et~al.(2023)Lam, Sanchez-Gonzalez, Willson, Wirnsberger, Fortunato, Alet, Ravuri, Ewalds, Eaton-Rosen, Hu, et~al.]{lam2023learning}
Remi Lam, Alvaro Sanchez-Gonzalez, Matthew Willson, Peter Wirnsberger, Meire Fortunato, Ferran Alet, Suman Ravuri, Timo Ewalds, Zach Eaton-Rosen, Weihua Hu, et~al.
\newblock Learning skillful medium-range global weather forecasting.
\newblock \emph{Science}, 382\penalty0 (6677):\penalty0 1416--1421, 2023.

\bibitem[Oono and Suzuki(2019)]{oono2019graph}
Kenta Oono and Taiji Suzuki.
\newblock Graph neural networks exponentially lose expressive power for node classification.
\newblock \emph{arXiv preprint arXiv:1905.10947}, 2019.

\bibitem[Alon and Yahav(2020)]{alon2020bottleneck}
Uri Alon and Eran Yahav.
\newblock On the bottleneck of graph neural networks and its practical implications.
\newblock \emph{arXiv preprint arXiv:2006.05205}, 2020.

\bibitem[Dwivedi and Bresson(2020)]{dwivedi2020generalization}
Vijay~Prakash Dwivedi and Xavier Bresson.
\newblock A generalization of transformer networks to graphs.
\newblock \emph{arXiv preprint arXiv:2012.09699}, 2020.

\bibitem[Ying et~al.(2021)Ying, Cai, Luo, Zheng, Ke, He, Shen, and Liu]{ying2021transformers}
Chengxuan Ying, Tianle Cai, Shengjie Luo, Shuxin Zheng, Guolin Ke, Di~He, Yanming Shen, and Tie-Yan Liu.
\newblock Do transformers really perform badly for graph representation?
\newblock \emph{Advances in neural information processing systems}, 34:\penalty0 28877--28888, 2021.

\bibitem[Mialon et~al.(2021)Mialon, Chen, Selosse, and Mairal]{mialon2021graphit}
Gr{\'e}goire Mialon, Dexiong Chen, Margot Selosse, and Julien Mairal.
\newblock Graphit: Encoding graph structure in transformers.
\newblock \emph{arXiv preprint arXiv:2106.05667}, 2021.

\bibitem[Ramp{\'a}{\v{s}}ek et~al.(2022)Ramp{\'a}{\v{s}}ek, Galkin, Dwivedi, Luu, Wolf, and Beaini]{rampavsek2022recipe}
Ladislav Ramp{\'a}{\v{s}}ek, Michael Galkin, Vijay~Prakash Dwivedi, Anh~Tuan Luu, Guy Wolf, and Dominique Beaini.
\newblock Recipe for a general, powerful, scalable graph transformer.
\newblock \emph{Advances in Neural Information Processing Systems}, 35:\penalty0 14501--14515, 2022.

\bibitem[Shirzad et~al.(2023)Shirzad, Velingker, Venkatachalam, Sutherland, and Sinop]{shirzad2023exphormer}
Hamed Shirzad, Ameya Velingker, Balaji Venkatachalam, Danica~J Sutherland, and Ali~Kemal Sinop.
\newblock Exphormer: Sparse transformers for graphs.
\newblock In \emph{International Conference on Machine Learning}, pages 31613--31632. PMLR, 2023.

\bibitem[Chalapathi et~al.(2024)Chalapathi, Du, and Krishnapriyan]{chalapathi2024scaling}
Nithin Chalapathi, Yiheng Du, and Aditi Krishnapriyan.
\newblock Scaling physics-informed hard constraints with mixture-of-experts.
\newblock \emph{arXiv preprint arXiv:2402.13412}, 2024.

\bibitem[Brody et~al.(2021)Brody, Alon, and Yahav]{brody2021attentive}
Shaked Brody, Uri Alon, and Eran Yahav.
\newblock How attentive are graph attention networks?
\newblock \emph{arXiv preprint arXiv:2105.14491}, 2021.

\bibitem[Li et~al.(2022{\natexlab{b}})Li, Huang, Liu, and Anandkumar]{li2022fourier}
Zongyi Li, Daniel~Zhengyu Huang, Burigede Liu, and Anima Anandkumar.
\newblock Fourier neural operator with learned deformations for pdes on general geometries.
\newblock \emph{arXiv preprint arXiv:2207.05209}, 2022{\natexlab{b}}.

\bibitem[Jin et~al.(2022)Jin, Meng, and Lu]{jin2022mionet}
Pengzhan Jin, Shuai Meng, and Lu~Lu.
\newblock Mionet: Learning multiple-input operators via tensor product.
\newblock \emph{arXiv preprint arXiv:2202.06137}, 2022.

\bibitem[Qi et~al.(2017)Qi, Yi, Su, and Guibas]{qi2017pointnet++}
Charles~Ruizhongtai Qi, Li~Yi, Hao Su, and Leonidas~J Guibas.
\newblock Pointnet++: Deep hierarchical feature learning on point sets in a metric space.
\newblock \emph{Advances in neural information processing systems}, 30, 2017.

\bibitem[Farin(2014)]{farin2014curves}
Gerald Farin.
\newblock \emph{Curves and surfaces for computer-aided geometric design: a practical guide}.
\newblock Elsevier, 2014.

\end{thebibliography}
\appendix
\section{Datasets and Model Hyperparameters.}
\label{Appendix:dataset}
\subsection{Datasets}

\textbf{NS.} 
We use a two-dimensional steady-state fluid dynamics dataset governed by the incompressible Navier–Stokes equations. The computational domain is a square region of size $[0, 8]^2$ with four internal circular cavities, resulting in a complex and non-trivial geometry. The goal is to predict the velocity components in the $x$ and $y$ directions, denoted by $u$ and $v$, respectively, along with the pressure field $p$, given the input mesh geometry. The domain is defined as
\begin{align}
\Omega = [0, 8]^2 \setminus \bigcup_{i = 1}^4 R_i
\end{align}
where each $R_i$ represents a circular cavity. The flow within this domain is governed by the steady-state incompressible Navier–Stokes equations:
\begin{align}
(\mathbf{u} \cdot \nabla) \mathbf{u} &= \frac{1}{\mathrm{Re}} \nabla^2 \mathbf{u} - \nabla p \\
\nabla \cdot \mathbf{u} &= 0
\end{align}
where $\mathbf{u} = (u, v)$ is the velocity vector field and $\mathrm{Re}$ is the Reynolds number. The boundary conditions are defined as follows: the velocity is set to zero on the entire boundary, i.e., $\mathbf{u} = 0$ on $\partial \Omega$. A parabolic velocity profile is imposed at the inlet (left boundary), given by $u_x = \frac{y(8 - y)}{16}$, while at the outlet (right boundary), the pressure is fixed at $p = 0$. Each sample in the dataset consists of a two-dimensional unstructured mesh over the domain $\Omega$ and the corresponding geometric configuration determined by the locations of the circular cavities. The outputs are the velocity field $(u, v)$ and the pressure field $p$, evaluated over the mesh nodes. We use the dataset provided by \citet{hao2023gnot}, which is publicly available at \url{https://github.com/HaoZhongkai/GNOT}. It consists of 1,100 samples generated by varying the positions of the four cavities to create different internal geometries. Of these, 1,000 samples are used for training and 100 for testing.
\begin{figure}[htbp]
    \centering
    \includegraphics[width=0.8\textwidth]{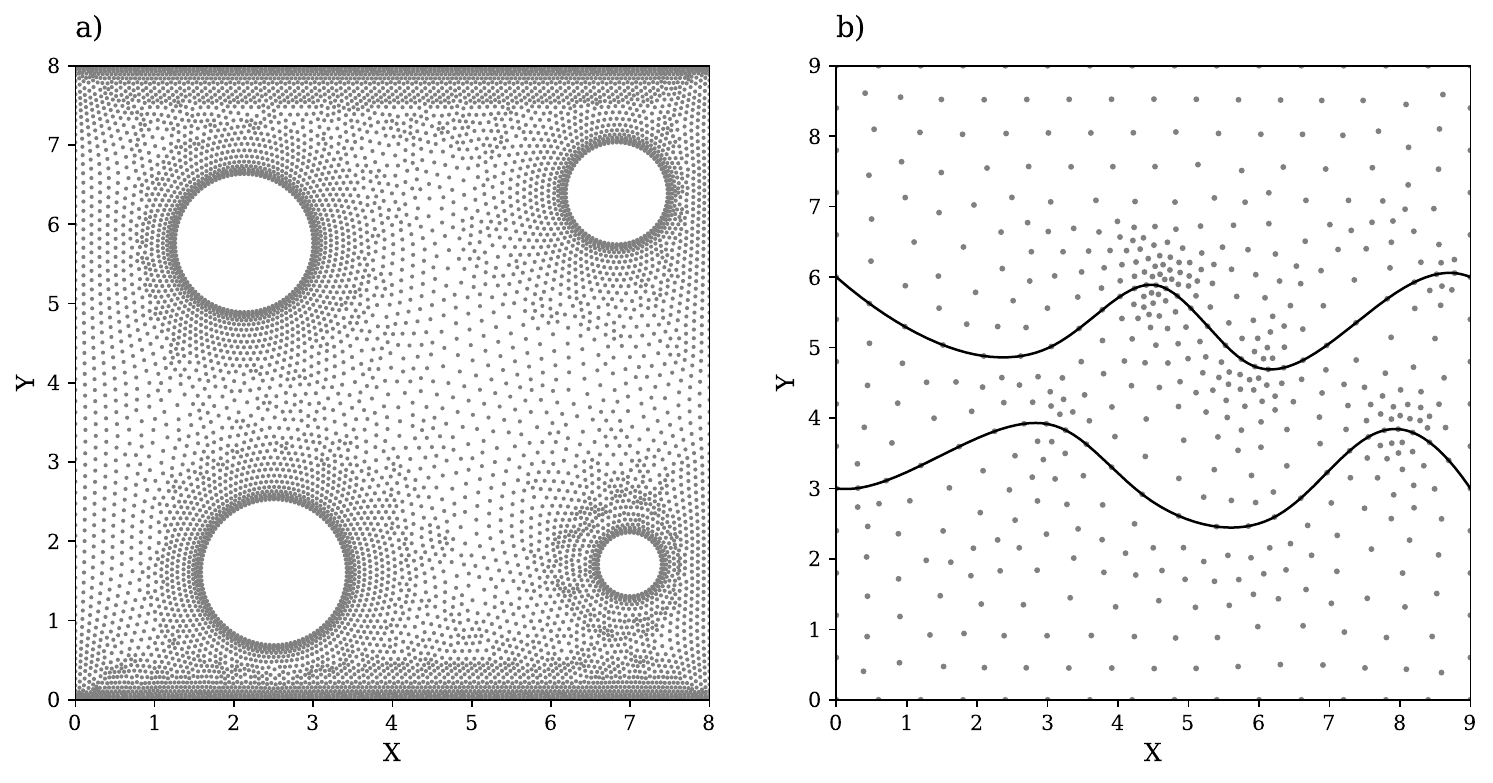}
    \caption{Visualization of mesh points depicting the domain geometry of a single sample from a) the NS dataset and b) the Heat dataset, where the geometry varies across samples.}
    \label{fig:mesh}
\end{figure}

\begin{figure}[htbp]
    \centering
    \includegraphics[width=0.8\textwidth]{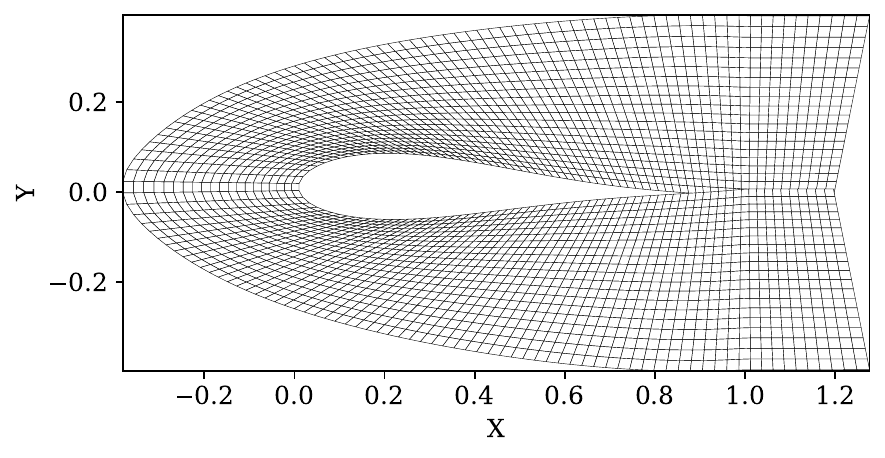}
    \caption{Visualization of mesh points depicting the domain geometry of a single sample from the Airfoil dataset}
    \label{fig:airfoilmesh}
\end{figure}

\textbf{Heat.}
We use a multi-scale heat conduction dataset, where the goal is to predict the steady-state temperature field $T$ given spatially varying material properties and complex boundary conditions. The problem is governed by the two-dimensional steady-state heat equation:
\begin{align}
\rho C_p \mathbf{u} \cdot \nabla T - k \nabla^2 T = Q
\end{align}
where $\rho$ is the density, $C_p$ is the specific heat capacity, $\mathbf{u}$ is the velocity field, $k$ is the thermal conductivity, and $Q$ is the internal heat source. The computational domain is a rectangle defined as $\Omega = [0, 9]^2$, segmented into three subdomains using two spline-shaped curves. Each subdomain possesses distinct physical properties, resulting in a highly heterogeneous medium. Periodic boundary conditions are applied along the left and right edges of the domain. The top boundary is assigned a temperature profile, which acts as an input function. Another input function parameterizes the shapes of the internal splines, thereby influencing the geometry of the subdomains and the overall distribution of material properties. Each sample in the dataset includes a two-dimensional unstructured mesh over the domain $\Omega$ and five input functions: the geometric configuration defined by the spline parameters and the temperature distribution prescribed on the top boundary. The output is the temperature field $T$ evaluated over the same discretized mesh. We use the dataset provided by \citet{hao2023gnot}, which is publicly available at \url{https://github.com/HaoZhongkai/GNOT}. It consists of 1,100 samples generated by varying the spline parameters and boundary conditions to produce diverse thermal configurations. Of these, 1,000 samples are used for training and 100 for testing.

\textbf{Airfoil.}
We examine transonic flow over an airfoil, governed by the inviscid Euler equations: 
\begin{align}
\frac{\partial \rho^f}{\partial t} + \nabla \cdot (\rho^f \bm{v}) = 0, \quad \quad
\frac{\partial \rho^f\bm{v}}{\partial t} + \nabla \cdot (\rho^f \bm{v} \otimes \bm{v} + p \mathbf{I}) = 0, \quad \quad
\frac{\partial E}{\partial t} + \nabla \cdot  \Bigl( (E + p)\bm{v} \Bigr) = 0 
\end{align}

where $\rho^f$ denotes fluid density, $\bm{v}$ is the velocity field, $p$ is the pressure, and $E$ represents total energy. Viscous effects are neglected in this formulation. The far-field boundary conditions are specified as: $\rho_{\infty} = 1$, $p_{\infty} = 1.0$, $M_{\infty} = 0.8$, and angle of attack ($AoA$) equal to 0°. On the airfoil surface, a no-penetration condition is enforced. To represent the airfoil geometry, we use the design element method provided in \citet{farin2014curves}. Starting from the baseline NACA-0012 profile, the geometry is embedded into a cubic design element framework comprising eight control nodes. Shape deformation is achieved by displacing these control nodes vertically, with the displacements sampled from a uniform distribution $d \sim \mathbb{U}[-0.05, 0.05]$. Each sample in the dataset consists of a two-dimensional structured mesh of size \(221 \times 51\), along with a single input function representing the airfoil geometry defined by 8 control nodes. The output is the Mach number \(M\), evaluated at each point on the same discretized mesh. We use the dataset provided by \citet{li2022fourier}, which is publicly available at \url{https://github.com/neuraloperator/Geo-FNO}. The dataset contains over 2,000 samples generated from various airfoil shapes. Out of these, 1,000 samples are used for training and 200 for testing.

\subsection{Model Hyperparameters}
Table~\ref{tab:hyperparameters} outlines the hyperparameters used in our experiments for all benchmarks across the GNOT and GITO models. To ensure a fair comparison, we kept the number of attention layers, MLP layers, and experts identical across both models. However, since GITO employs a fusion layer that concatenates the outputs of the GNN and self-attention components within the HGT module (effectively doubling the hidden dimension) we set the hidden size in GNOT to be twice that of GITO prior to this fusion. Note that the hidden sizes reported in the table for GITO correspond to the dimensions before concatenation. For the Heat dataset, which involves larger inputs and more complex dynamics, we increased the number of MLP layers and experts in GNOT to match or slightly exceed the total parameter count of GITO. Although exact parameter matching was not possible due to architectural differences, we intentionally allocated slightly more parameters to GNOT to fairly showcase its performance and demonstrate its modeling efficacy under comparable capacity constraints. All experiments were conducted on a single NVIDIA V100 GPU with 8 vCPUs and 52 GB of RAM.
\begin{table}[h]
\centering
\begin{tabular}{@{}c|cc|cc|cc@{}}
\toprule
\multirow{2}{*}{\textbf{Hyperparameter Type}} &
  \multicolumn{2}{c|}{\textbf{NS}} &
  \multicolumn{2}{c|}{\textbf{Heat}} &
  \multicolumn{2}{c}{\textbf{Airfoil}} \\ \cmidrule(l){2-7} 
 &
  \multicolumn{1}{c|}{\textbf{GNOT}} &
  \textbf{GITO} &
  \multicolumn{1}{c|}{\textbf{GNOT}} &
  \textbf{GITO} &
  \multicolumn{1}{c|}{\textbf{GNOT}} &
  \textbf{GITO} \\ \midrule
Activation function            & \multicolumn{1}{c|}{GELU} & GELU & \multicolumn{1}{c|}{GELU} & GELU & \multicolumn{1}{c|}{GELU} & GELU \\
Number of attention layers     & \multicolumn{1}{c|}{2}    & 2    & \multicolumn{1}{c|}{3}    & 3    & \multicolumn{1}{c|}{2}    & 2    \\
Hidden size of attention       & \multicolumn{1}{c|}{192}  & 96   & \multicolumn{1}{c|}{256}  & 128  & \multicolumn{1}{c|}{192}  & 96   \\
Layers of MLP                  & \multicolumn{1}{c|}{2}    & 2    & \multicolumn{1}{c|}{4}    & 3    & \multicolumn{1}{c|}{2}    & 2    \\
Hidden size of MLP             & \multicolumn{1}{c|}{192}  & 96   & \multicolumn{1}{c|}{256}  & 128  & \multicolumn{1}{c|}{192}  & 96   \\
Hidden size of input embedding & \multicolumn{1}{c|}{192}  & 96   & \multicolumn{1}{c|}{256}  & 128  & \multicolumn{1}{c|}{192}  & 96   \\
Learning rate schedule &
  \multicolumn{1}{c|}{OneCycle} &
  OneCycle &
  \multicolumn{1}{c|}{OneCycle} &
  OneCycle &
  \multicolumn{1}{c|}{OneCycle} &
  OneCycle \\
N experts                      & \multicolumn{1}{c|}{2}    & 2    & \multicolumn{1}{c|}{4}    & 3    & \multicolumn{1}{c|}{2}    & 2    \\
N heads                        & \multicolumn{1}{c|}{8}    & 8    & \multicolumn{1}{c|}{8}    & 8    & \multicolumn{1}{c|}{8}    & 8    \\
Total model parameters &
  \multicolumn{1}{c|}{4.49M} &
  4.37M &
  \multicolumn{1}{c|}{21.26M} &
  18.24M &
  \multicolumn{1}{c|}{4.48M} &
  4.37M \\ \bottomrule
\end{tabular}
\caption{Hyperparameters and training runtime details for the GNOT and GITO models on the NS, Heat and Airfoil datasets. Hidden sizes for GITO are reported before fusion.}
\label{tab:hyperparameters}
\end{table}






\end{document}